\documentclass[conference]{IEEEtran}
\IEEEoverridecommandlockouts

\usepackage{cite}
\usepackage{amsmath,amssymb,amsfonts,hyperref}
\usepackage{algorithmic,booktabs,multirow}
\usepackage{graphicx}
\usepackage{textcomp}
\usepackage{xcolor}
\def\BibTeX{{\rm B\kern-.05em{\sc i\kern-.025em b}\kern-.08em
    T\kern-.1667em\lower.7ex\hbox{E}\kern-.125emX}}
\begin{document}

\title{DACAT: Dual-stream Adaptive Clip-aware Time Modeling for Robust Online Surgical Phase Recognition
}

\author{
\IEEEauthorblockN{Kaixiang Yang$^{a}$,
Qiang Li$^{a,*}$, Zhiwei Wang$^{a,*}$}
\IEEEauthorblockA{\textit{$^a$Wuhan National Laboratory for Optoelectronics, Huazhong University of Science and Technology} \\
*Corresponding authors: \{liqiang8, zwwang\}@hust.edu.cn}
}



\maketitle

\begin{abstract}
Surgical phase recognition has become a crucial requirement in laparoscopic surgery, enabling various clinical applications like surgical risk forecasting.
Current methods typically identify the surgical phase using individual frame-wise embeddings as the fundamental unit for time modeling. However, this approach is overly sensitive to current observations, often resulting in discontinuous and erroneous predictions within a complete surgical phase.
In this paper, we propose DACAT, a novel dual-stream model that adaptively learns clip-aware context information to enhance the temporal relationship. In one stream, DACAT pretrains a frame encoder, caching all historical frame-wise features. In the other stream, DACAT fine-tunes a new frame encoder to extract the frame-wise feature at the current moment. Additionally, a max clip-response read-out (Max-R) module is introduced to bridge the two streams by using the current frame-wise feature to adaptively fetch the most relevant past clip from the feature cache. The clip-aware context feature is then encoded via cross-attention between the current frame and its fetched adaptive clip, and further utilized to enhance the time modeling for accurate online surgical phase recognition.
The benchmark results on three public datasets, \textit{i.e.}, Cholec80, M2CAI16, and AutoLaparo, demonstrate the superiority of our proposed DACAT over existing state-of-the-art methods, with improvements in Jaccard scores of at least 4.5\%, 4.6\%, and 2.7\%, respectively. Our code and models have been released at {\href{https://github.com/kk42yy/DACAT}{https://github.com/kk42yy/DACAT}}.
\end{abstract}


\section{Introduction}
Minimally invasive surgery must adhere to specific procedures and steps to ensure maximum safety and efficiency. Automatic online recognition of surgical phases is a critical component in surgical workflow analysis. This technology enables the forecasting of surgical risks and allows for the timely preparation of necessary tools.
However, online phase recognition faces significant challenges, including irrelevant frames caused by camera movement, interference from blood and smoke, and variations in surgical phase sequences due to individual differences among patients. 

With the advent of deep learning, surgical phase recognition has made significant progress.
Existing deep learning-based solutions typically employ a frame encoder, such as ResNet~\cite{ref10resnet} or Transformer~\cite{ref6Transformer,ref7ViT}, to extract individual frame-wise embeddings, followed by a temporal modeling module, such as Long Short-Term Memory (LSTM)~\cite{ref4LSTM} or Temporal Convolutional Networks (TCN)~\cite{ref5TCN}, to make the final decisions based on the spatio-temporal features.
These approaches can be categorized into multi-task and single-task learning.


The multi-task learning methods~\cite{ref11multi,ref12multi,ref13multimtrcnet} rely on additional annotations, such as surgical tool presence, to enhance the recognition of surgical phases. This approach increases both labor and computational burdens.
In contrast, single-task learning methods~\cite{ref14lstmSV-RCNet,ref15lstm,ref16lstm,ref17lstmTMRNet,ref18lstmBNPitfalls,ref19TCNTECNO,ref20TransOpera,ref21TransTransSVNet,ref22Translovit,ref23TransCMTNet,ref24TransSkit} focus solely on surgical phase recognition, requiring only frame-level annotations for training. For example, the most recent work~\cite{ref18lstmBNPitfalls} proved that using Batch Normalization~\cite{ref25BN} free backbones, and carrying the hidden state of LSTM can significantly improve the performance.
Despite variations in architectural designs, these methods generally follow a common paradigm: encoding individual frame-wise embeddings and then modeling their temporal relationships. However, frame-wise embeddings are vulnerable to various types of interference, such as blood stains or smoke, making the temporal relationships sensitive to the current frame's observations. Additionally, individual differences among patients can cause variations in the surgical phase sequence, exacerbating the impact of these inaccurate spatial features on subsequent temporal processing.

In this paper, instead of solely relying on unreliable frame-wise embeddings, we propose DACAT, a Dual-stream network that encodes temporal context information as Adaptive Clip-aware embeddings, facilitating the learning of robust spatio-temporal representations for online surgical phase recognition.
Our main contributions are summarized as follows:
\begin{itemize}
    \item We propose a novel DACAT for online surgical phase recognition, which combine dual-stream model that adaptively learns clip-aware context information to enhance the temporal relationship. Adaptive clip helps to filter out irrelevant interference of current frame, thereby accurately identifying the phase. 
    \item We design a parameter-free max clip-response read-out (Max-R) mechanism to retrieve the most relevant clip for the current frame. To avoid reduplicated encoding for each frame, we also implement a feature cache, and use cross-attention mechanism to integrate cached and current features, thereby improving the inference speed of the algorithm to meet the demands of clinical applications.
    \item DACAT outperforms the state-of-the-art (SOTA) methods on three datasets, Cholec80, M2CAI16, and AutoLaparo, across two types of surgeries, achieving improvement of 4.5\%, 4.6\%, and 2.7\% in Jaccard, which demonstrates the superiority of our proposed method.
\end{itemize}

\section{Methodology}
\label{sec:method}

\begin{figure}
\centerline{\includegraphics[width=\columnwidth]{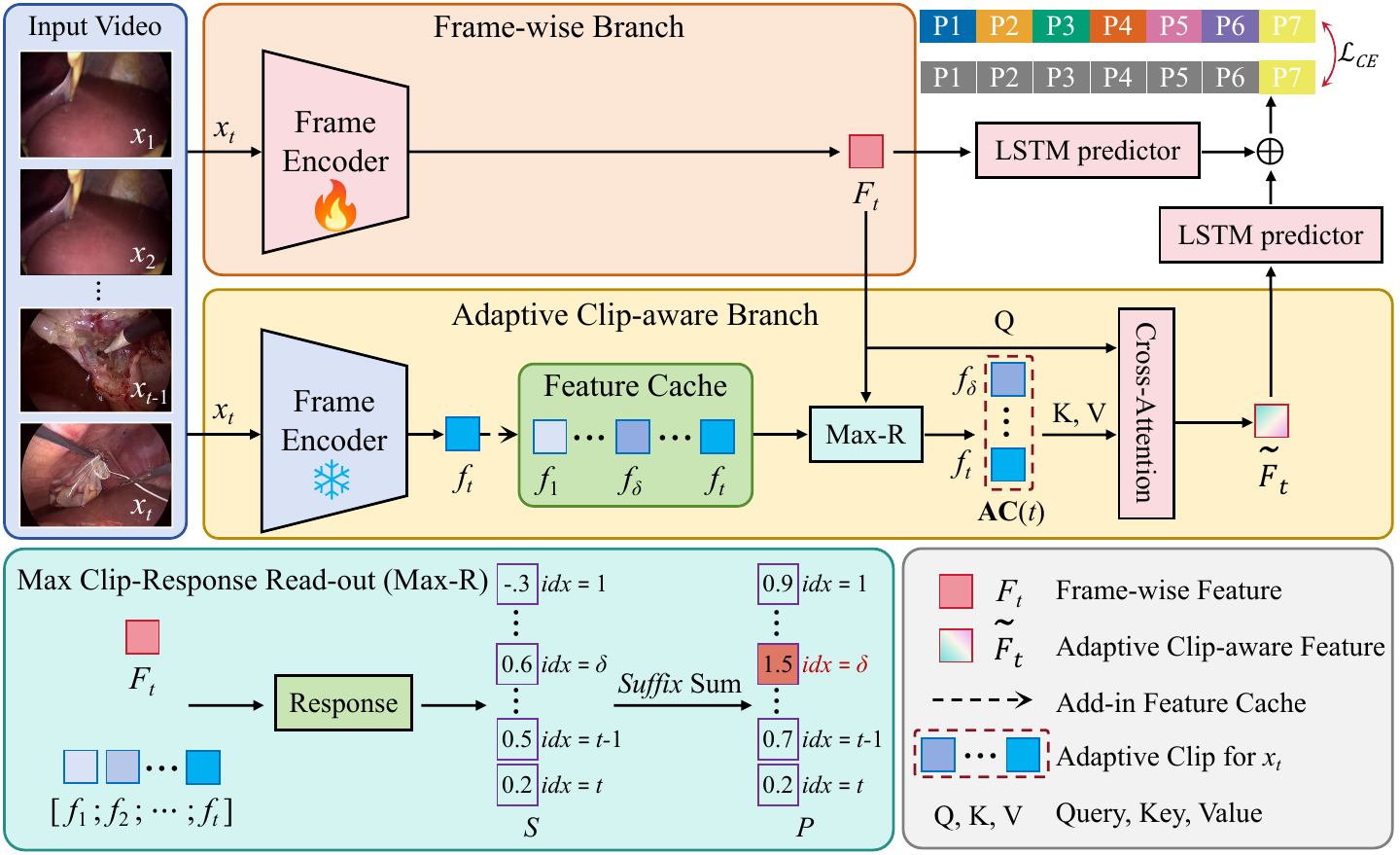}}
\vspace{-5pt}
\caption{The overall framework of DACAT, which consists of two main branches, (1) Frame-wise Branch (FWB) and (2) Adaptive Clip-aware Branch (ACB). FWB extracts the embeddings for current frame $x_t$. ACB obtains the most relevant clip-aware features for $x_t$ through designed Max Clip-Response Read-out (Max-R) and cross-attention (CA). Finally, combining the results of FWB and ACB to obtain the phase prediction. $S$, $P$, and $\mathbf{AC}(t)$ are frame response matrix, clip response matrix and adaptive clip, respectively. Response is formulated as Eq.~\ref{response_eq}.} 
\vspace{-15pt}
\label{fig2_method}
\end{figure}

Fig.~\ref{fig2_method} shows the framework of our proposed DACAT. It consists of two main branches, \textit{i.e.}, (i) Frame-wise Branch (FWB) processing the frame-wise feature and (ii) Adaptive Clip-aware Branch (ACB) which reads out the most relevant clip with the current frame from pre-trained feature cache and integrates these frame-wise features into adaptive clip-aware feature through cross-attention (CA) module. DACAT enhances the relevant context and filter out interference for current frame, which reduces the the complexity of temporal processing and leads to more accurate phase identification.

\subsection{Frame-wise Branch}

\cite{ref18lstmBNPitfalls} utilizes structure without Batch Normalization like ConvNeXt~\cite{ref26convnext} and LSTM for phase recognition, and achieves the SOTA performance in phase recognition. We modify the network structure and adopt the training skills as our FWB, which consists of (1) using ConvNeXt V2-T~\cite{ref27convnextv2} as our frame encoder, (2) freezing the bottom layers (\textit{i.e.}, Block 1, 2, 3) of the encoder and (3) carrying the hidden state in LSTM to enlarge the temporal field. Through the frame encoder, \textbf{Frame-wise feature} $F_t$ is obtained for the current frame $x_t$. In the temporal feature procedure, carrying hidden state from processing $x_1$ to $x_t$ will expand the temporal perspective to provide the model with information from the start to time $t$, enabling phase identification on long-term scale.

\subsection{Feature Cache}

Most video object segmentation (VOS) methods employ a feature memory~\cite{ref28spacememory} to store useful segmentation features for subsequent frame, which enables the model to retrieve previously acquired information. 
Drawing inspiration from this, we construct a feature cache to store encoded features, which also avoids repeated frame encoding.
Additionally, phase recognition task does not require excessive spatial information, our cache only retain the compressed single-frame features that have undergone average pooling. At time $t$, the features stored in cache can be formulate as:
\begin{equation}
    \mathrm{Cache}(t) = \{f_1, f_2, \cdots, f_t\}, f_i \in \mathbb{R}^d
\end{equation}
where $d$ is the feature dimension after encoder and set to $768$.
Consequently, ours can accommodate spatially compressed features for all frames from the beginning of a video.

To ensure the cache provides accurate and useful features, we fine-tune ConvNeXt V2-T with its pre-trained parameters using the Frame-wise Branch and LSTM predictor. We also explore whether to use fine-tuning or real-time training for feature cache, with more details in \hyperlink{pretrain}{Sec III.C}.

\subsection{Adaptive Clip-aware Branch}

Frame-wise feature based phase recognition is highly sensitive to current observations, which can introduce interference and noise into subsequent temporal processing, leading to discontinuous and erroneous prediction within a surgical phase. Furthermore, using excessive or insufficient frame information can also exacerbate the difficulty of phase recognition. 
Therefore, we design Adaptive Clip-aware Branch (ACB) to obtain adaptive clip-aware feature for current frame, which mitigates the impact of interference on recognition.

Initially, we propose a parameter-free max clip-response read-out (Max-R) module to retrieve the most relevant clip for each frame from feature cache, \textit{i.e.}, adaptive clip. 
As shown in Fig.~\ref{fig2_method}, we utilize the extracted frame $F_t$ in FWB to search for its adaptive clip $\mathbf{AC}(t)$ in feature cache. 
It is necessary to assess the correlation between the current and preceding frames. Considering that the features stored in cache are encoded by pre-trained extractor and reliable, and that the frame encoder in ACB also employs above fine-tuned parameters, we calculate the correlation using direct multiplication between $F_t$ and $\{f_i\}^t_{i=1}$, where larger product indicates higher correlation.
The frame response matrix $S$ is:
\begin{equation}\label{response_eq}
    S = F_t^{\mathrm{T}} * 
    \begin{pmatrix}
        f_1; f_2; \cdots; f_t
    \end{pmatrix}
    \in \mathbb{R}^{1\times t}
\end{equation}
where $S_i$ represents the response between $x_t$ and $x_i$. 
Subsequently, to measure the overall correlation of different clip, we compute the \textit{suffix} sum for $S$ to obtain clip response matrix $P$, where larger value of $P(j)$ indicates higher correlation for the clip from time $j$ to $t$. $P$ is formulated as:
\begin{equation}
    P = 
    \begin{pmatrix}
        \displaystyle\sum_{j=1}^{t}S_j; \displaystyle\sum_{j=2}^{t}S_j;\cdots;
        \displaystyle\sum_{j=t-1}^{t}S_j; \displaystyle\sum_{j=t}^{t}S_j
    \end{pmatrix}
    \in \mathbb{R}^{1\times t}
\end{equation}
and adaptive clip can be generated as:
\begin{equation}
    \mathbf{AC}(t)=
    \begin{pmatrix}
        f_\delta; \cdots; f_{t-1}; f_t
    \end{pmatrix}
\end{equation}
where $\delta = \underset{j}{\operatorname{argmax}} \ P(j)$.

In order to make better use of the information of $\mathbf{AC}(t)$ and simultaneously reduce the gap between two branches, we employ CA to integrate $\mathbf{AC}(t)$ into $F_t$ to obtain the \textbf{Adaptive Clip-aware feature} $\tilde{F}_t$, thereby avoiding repeated LSTM temporal processing.
We utilize LSTM and Linear to make the phase probability of FWB and ACB, and combine them to obtain the final prediction. 

\subsection{Training Details}
We use Cross-Entropy Loss $\mathcal{L}_{CE}$ as our object function.
Our all experiments are conducted on a single NVIDIA GeForce RTX 4090 24GB GPU. We resize the images to $216\times 384$ pixels. We train the feature cache extractor using AdamW with a learning rate $1e-4$ (AutoLaparo with $5e-4$), weight decay of $0.01$ for 200 epochs. A batch size of 1 and input segment length of 256 are set. The DACAT is trained for 30 epochs using a batch size of 1 and input segment length of 64. AdamW optimizer with an initial learning rate of $1e-5$ and weight decay of $0.01$ is utilized for optimization. In inference, surgical phases are predicted frame by frame. 
Our code and models have been released at {\href{https://github.com/kk42yy/DACAT}{https://github.com/kk42yy/DACAT}}.


\begin{table}[t]
\caption{The comparison results (\%) with SOTA on the Cholec80, M2CAI16 and AutoLaparo.}
\vspace{-5pt}
\label{tab1_sota}
    \centering
    \resizebox{\linewidth}{!}{
    \begin{tabular}{c|c|cccc} \specialrule{1.5pt}{0pt}{0pt}
    Dataset & Methods & Accuracy & Precision & Recall & Jaccard \\ \hline
    \multirow{10}{*}{Cholec80} & SV-RCNet~\cite{ref14lstmSV-RCNet} & $85.3 \pm 7.3$ & $80.7 \pm 7.0$ & $83.5 \pm 7.5$ & $-$  \\
    & MTRCNet-CL*~\cite{ref13multimtrcnet} & $89.2\pm 7.6$ & $86.9\pm4.3$ & $88.0\pm6.9$ & $-$ \\
    & TMRNet~\cite{ref17lstmTMRNet} & $90.1 \pm 7.6$ & $90.3 \pm 3.3$ & $89.5 \pm 5.0$ & $79.1 \pm 5.7$ \\
    & Trans-SVNet~\cite{ref21TransTransSVNet} & $90.3 \pm 7.1$ & $90.7 \pm 5.0$ & $88.8 \pm 7.4$ & $79.3 \pm 6.6$ \\
    & UATD~\cite{ref32uatd} & $91.9 \pm 5.6$ & $89.5 \pm 4.4$ & $90.5 \pm 5.9$ & $79.9 \pm 8.5$ \\
    & CMTNet~\cite{ref23TransCMTNet} & $92.9 \pm 5.9$ & $90.1 \pm 7.1$ & $92.0 \pm 4.4$ & $81.5 \pm 10.4$ \\
    & LoViT~\cite{ref22Translovit} & $92.4 \pm 6.3$ & $89.9 \pm 6.1$ & $90.6 \pm 4.4$ & $81.2 \pm 9.1$ \\
    & SKiT~\cite{ref24TransSkit} & $93.4 \pm 5.2$ & 90.9 & 91.8 & 82.6 \\
    & BNpitfalls~\cite{ref18lstmBNPitfalls} & $93.5 \pm 6.5$ & \underline{$91.6 \pm 5.0$} & \underline{$91.4 \pm 9.3$} & $82.9 \pm 10.1$  \\
    & DACAT (Ours) & $\mathbf{95.5 \pm 4.3}$ & $\mathbf{93.6 \pm 4.1}$ & $\mathbf{93.4 \pm 5.3}$ & $\mathbf{87.4 \pm 8.1}$ \\
    \hline

    \multirow{6}{*}{M2CAI16} & SV-RCNet~\cite{ref14lstmSV-RCNet} & $81.7 \pm 8.1$ & $81.0 \pm 8.3$ & $81.6 \pm 7.2$ & $65.4 \pm 8.9$ \\
    & TMRNet~\cite{ref17lstmTMRNet} & $87.0 \pm 8.6$ & $87.8 \pm 6.9$ & $88.4 \pm 5.3$ & $75.1 \pm 6.9$ \\
    & Trans-SVNet~\cite{ref21TransTransSVNet} & $87.2 \pm 9.3$ & $88.0 \pm 6.7$ & $87.5 \pm 5.5$ & $74.7 \pm 7.7$ \\
    & UATD~\cite{ref32uatd} & $87.6 \pm 8.7$ & $88.2 \pm 7.4$ & $87.9 \pm 9.6$ & $75.7 \pm 9.5$ \\
    & CMTNet~\cite{ref23TransCMTNet} & $88.2 \pm 9.2$ & $88.3 \pm 7.8$ & $88.7 \pm 6.2$ & $76.1 \pm 9.2$ \\
    & DACAT (Ours) & $\mathbf{91.3 \pm 9.3}$ & $\mathbf{90.8 \pm 7.6}$ & $\mathbf{90.6 \pm 6.7}$ & $\mathbf{80.7 \pm 8.8}$ \\
    \hline

    \multirow{8}{*}{AutoLaparo} & SV-RCNet~\cite{ref14lstmSV-RCNet} & $75.6$ & $64.0$ & $59.7$ & $47.2$ \\
    & TMRNet~\cite{ref17lstmTMRNet} & $78.2$ & $66.0$ & $61.5$ & $49.6$ \\
    & Trans-SVNet~\cite{ref21TransTransSVNet} & $78.3$ & $64.2$ & $62.1$ & $50.7$ \\
    & LoViT~\cite{ref22Translovit} & $81.4 \pm 7.6$ & $\mathbf{85.1}$ & $65.9$ & $55.9$ \\
    & SKiT~\cite{ref24TransSkit} & $82.9 \pm 6.8$ & $81.8$ & $70.1$ & $59.9$ \\
    & SPHASE~\cite{ref33sphase} & $83.8$ & $75.7$ & $71.3$ & $57.8$ \\
    & BNpitfalls~\cite{ref18lstmBNPitfalls} & $86.8 \pm 1.5$ & $78.2$ & $72.0$ & $64.2$  \\
    & DACAT (Ours) & $\mathbf{87.8 \pm 7.6}$ & $78.5$ & $\mathbf{75.0}$ & $\mathbf{66.9}$ \\
     
    \specialrule{1.5pt}{0pt}{0pt}
    \multicolumn{6}{l}{*: Multi-task learning methods.} \\
    \multicolumn{6}{l}{\underline{Underline}: Calculated by released weight.} \\
    \multicolumn{6}{l}{\textbf{Bold}: The best result.} \\
    \end{tabular}
    }
    \vspace{-20pt}
\end{table}

\section{Experiments}
\subsection{Materials and Metrics}

\textbf{Datasets.}
We have conducted extensive experiments on three datasets corresponding to two types of laparoscopic surgeries: Cholec80~\cite{ref29Cholec80} and M2CAI16~\cite{ref30m2cai16} for cholecystectomy, and AutoLaparo~\cite{ref31autolaparo} for hysterectomy. \textit{Cholec80} consists of 80 videos and seven phases. We follow the data split completely consistent with the previous works~\cite{ref24TransSkit,ref18lstmBNPitfalls}, 40 for training and the rest for testing. \textit{M2CAI16} contains 41 videos with eight phases. We split the dataset into 27 for training and 14 for testing following \cite{ref17lstmTMRNet,ref23TransCMTNet}. \textit{AutoLaparo} comprises 21 videos with seven phases. 10, 4 and 7 videos are spilt for training, validation and testing, respectively, following \cite{ref24TransSkit,ref18lstmBNPitfalls}. Three datasets are recorded at 25 frames per second (fps), and are sampled into 1 fps as \cite{ref24TransSkit,ref18lstmBNPitfalls}.

\textbf{Metrics.}
For quantitative evaluation, we employ four commonly-used metrics for phase recognition, \textit{i.e.}, accuracy, precision, recall and Jaccard. Accuracy is a video-based evaluation, while the rest are phase-based. For Cholec80 and M2CAI16, we follow the official challenge evaluation criteria~\cite{ref30m2cai16} and previous methods~\cite{ref17lstmTMRNet,ref24TransSkit,ref18lstmBNPitfalls}, considering a frame prediction to be correct if falling within 10 seconds of the ground truth.


\subsection{Comparison with the state-of-the-arts}

We compare our method with recent SOTA methods on three datasets, and the results are listed in Table~\ref{tab1_sota}. DACAT significantly outperforms previous SOTA methods across four evaluation metrics on all three datasets (except for the Precision on AutoLaparo). 
Specifically, on Cholec80, M2CAI16, and AutoLaparo, our method improves Accuracy and Jaccard by 2.0\% and 4.5\%, 3.1\% and 4.6\%, and 1.0\% and 2.7\%, respectively. 
These outstanding performance across different types of surgeries demonstrate the strong generalization ability of our approach. Additionally, during the inference phase, our per-frame online processing achieves speed of \textbf{38.1 fps}, higher than the video sampling rate of 25 fps, meeting the requirements for clinical applications.

\begin{figure}[t]
\centerline{\includegraphics[width=\columnwidth]{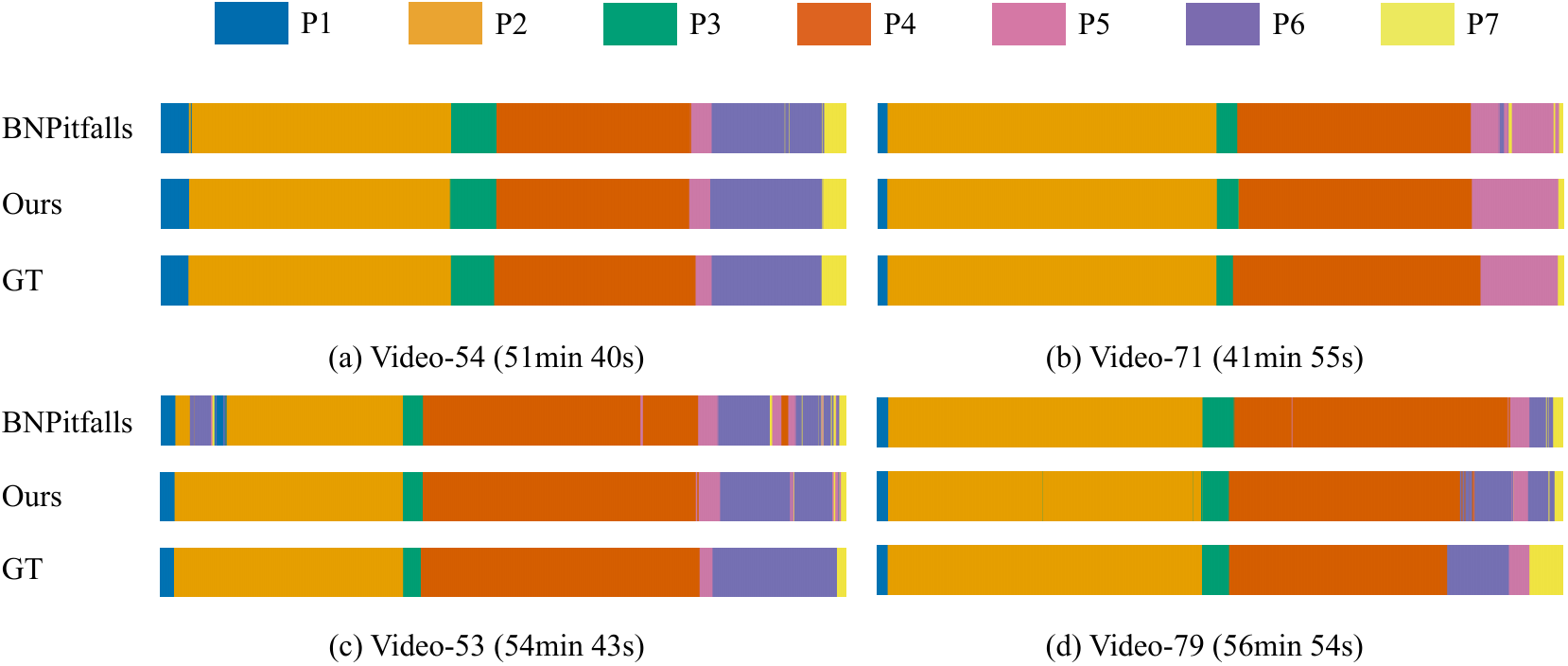}}
\vspace{-8pt}
\caption{Visualization comparison with previous SOTA on Cholec80. (a) and (b) represent good prediction, while (c) and (d) show relatively poor results.} 
\vspace{-10pt}
\label{fig3_visual}
\end{figure}

Fig.~\ref{fig3_visual} visualizes several testing cases. Compared with the latest SOTA, it is evident that our proposed combination of frame-wise and adaptive clip-aware features better handle complex surgical scenarios and obtain clearer phase boundaries and smoother inner-phase recognition. Overall, both the quantitative evaluation and qualitative visual comparisons clearly demonstrate the superiority of DACAT.

\subsection{Ablation Study}

\textbf{Effectiveness of Frame-wise Branch and Adaptive Clip-aware Branch.}
We first explore the roles of two branches in DACAT. As shown in Table~\ref{tab2_abaFWBetal}, only with the ACB also outperforms SOTA, indicating that adaptive clip-aware feature of each frame is able to provide suitable and less interfering clues. Combining with FWB, adaptive clip-aware feature is enhanced by frame-wise feature in spatial and temporal level, and also help filter out irrelevant information for FWB.
\vspace{-10pt}

\begin{table}[htbp]
\caption{Ablation study (\%) of two branches in DACAT on Cholec80.}
\vspace{-5pt}
\label{tab2_abaFWBetal}
    \centering
    \resizebox{\linewidth}{!}{
    \begin{tabular}{c|cccc} \specialrule{1.5pt}{0pt}{0pt}
    Methods &  Accuracy & Precision & Recall & Jaccard \\ \hline
    $w/o$ ACB & $94.1 \pm 5.7$ & $92.5 \pm 3.3$ & $91.3 \pm 9.1$ & $84.4 \pm 9.1$ \\
    $w/o$ FWB & $95.1 \pm 4.4$ & $93.2 \pm 4.7$ & $92.3 \pm 4.7$ & $85.5 \pm 10.5$ \\
    $both$ & $\mathbf{95.5 \pm 4.3}$ & $\mathbf{93.6 \pm 4.1}$ & $\mathbf{93.4 \pm 5.3}$ & $\mathbf{87.4 \pm 8.1}$ \\
    \specialrule{1.5pt}{0pt}{0pt}
    \end{tabular}
    }
    \vspace{-5pt}
\end{table}

\textbf{Fusion of FWB and ACB.}
In Table~\ref{tab3_abaFusion}, we then investigate the fusion of FWB and ACB, fusing $before$ and $after$ (DACAT) temporal processing. The $before$ obtains higher accuracy and precision, indicating that combining frame-wise and adaptive clip-aware feature improves temporal processing. Considering the overall performance, we ultimately choose to fuse after, allowing each branch to fully leverage their features.
\vspace{-15pt}

\begin{table}[htbp]
\caption{Ablation study (\%) of feature fusion in DACAT on Cholec80.}
\vspace{-5pt}
\label{tab3_abaFusion}
    \centering
    \resizebox{\linewidth}{!}{
    \begin{tabular}{c|cccc} \specialrule{1.5pt}{0pt}{0pt}
    Methods &  Accuracy & Precision & Recall & Jaccard \\ \hline
    $before$ & $\mathbf{95.7 \pm 3.5}$ & $\mathbf{94.0 \pm 4.7}$ & $92.4 \pm 7.2$ & $86.5 \pm 9.5$ \\
    $after$ & $95.5 \pm 4.3$ & $93.6 \pm 4.1$ & $\mathbf{93.4 \pm 5.3}$ & $\mathbf{87.4 \pm 8.1}$ \\
    \specialrule{1.5pt}{0pt}{0pt}
    \end{tabular}
    }
    \vspace{-8pt}
\end{table}

\begin{figure}[t]
\centerline{\includegraphics[width=\columnwidth]{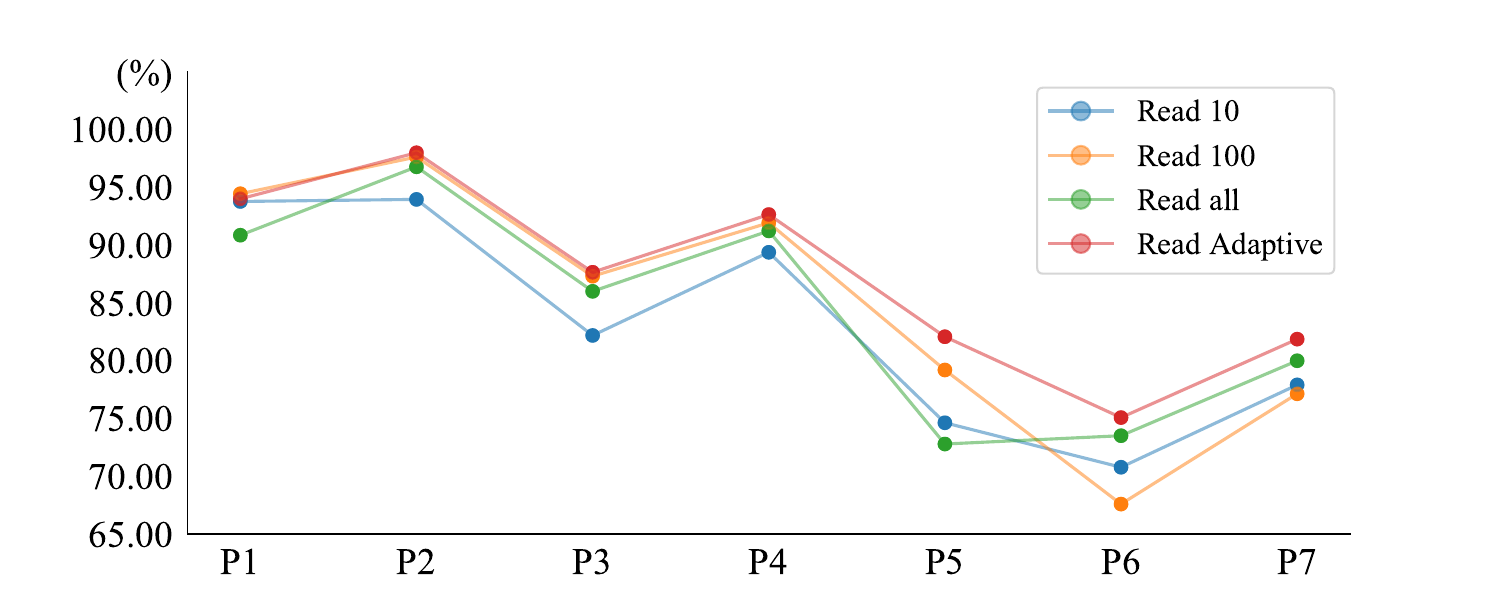}}
\vspace{-8pt}
\caption{The Jaccard of four read-out ways with respect to phase on Cholec80.} 
\vspace{-15pt}
\label{fig5_plot}
\end{figure}

\textbf{Read-out way from cache.}
Additionally, we research different Read-out ways from cache. In Fig~\ref{fig5_plot}, ``Read $10$" and ``Read $100$" represent read out features 
$\begin{pmatrix}
    f_{t-9}; \cdots; f_t
\end{pmatrix}$ and 
$\begin{pmatrix}
    f_{t-99}; \cdots; f_t
\end{pmatrix}$
for CA, respectively, while "Read all" refers to reading all features in cache at time $t$. ``Read Adaptive'' is our DACAT. 
Different read-out ways perform variably across phases. For instance, ``Read all'' achieves better Jaccard in P6 and P7, while ``Read 100'' performs well in P1 to P5. This indicates that different phases require specific information. Ours that uses adaptive clip, significantly outperforms the other three read-out ways across phases.

\hypertarget{hidden}{}
\textbf{Interaction way of $\mathbf{AC}(t)$ and $F_t$.}
We also investigate ways for obtaining adaptive clip-aware feature $\tilde{F}_t$, \textit{i.e.}, the integration between $\mathbf{AC}(t)$ and $F_t$. We explore three different approaches, with the first two rows in Table~\ref{tab3_abaMaxR} representing summing and concatenating two features. To maintain consistent feature dimensions, we first applied average pooling along the temporal dimension to $\mathbf{AC}(t)$. The results indicate that 
with CA (DACAT) obtains better integration.
\vspace{-10pt}

\begin{table}[htbp]
\caption{Ablation study (\%) of the interaction way between $\mathbf{AC}(t)$ and $F_t$ on Cholec80.}
\vspace{-5pt}
\label{tab3_abaMaxR}
    \centering
    \resizebox{\linewidth}{!}{
    \begin{tabular}{c|cccc} \specialrule{1.5pt}{0pt}{0pt}
    Methods &  Accuracy & Precision & Recall & Jaccard \\ \hline
    Add & $95.4 \pm 4.1$ & $92.6 \pm 5.8$ & $93.4 \pm 6.2$ & $86.2 \pm 10.1$ \\
    Concat & $95.4 \pm 4.6$ & $\mathbf{94.4 \pm 4.3}$ & $92.3 \pm 5.8$ & $86.9 \pm 8.0$ \\
    CA & $\mathbf{95.5 \pm 4.3}$ & $93.6 \pm 4.1$ & $\mathbf{93.4 \pm 5.3}$ & $\mathbf{87.4 \pm 8.1}$ \\
    \specialrule{1.5pt}{0pt}{0pt}
    \end{tabular}
    }
\end{table}

\hypertarget{pretrain}{}
\textbf{Effectiveness of fine-tuning for cache.}
As shown in Table~\ref{tab4_abaTraining}, we explore whether to fine-tune the frame encoder of cache. Without fine-tuning, we update the parameters when training the DACAT.
With fine-tuning produces better because fine-tuned cache provides reliable and stable features, and the absence of gradients simplifies the training process of ACB.

\begin{table}[htbp]
\caption{Ablation study (\%) of fine-tuning for cache on Cholec80.}
\vspace{-5pt}
\label{tab4_abaTraining}
    \centering
    \resizebox{\linewidth}{!}{
    \begin{tabular}{c|cccc} \specialrule{1.5pt}{0pt}{0pt}
    Methods &  Accuracy & Precision & Recall & Jaccard \\ \hline
    $w/o$ $fine$-$tuning$ & $94.7 \pm 4.3$ & $93.0 \pm 4.7$ & $91.8 \pm 8.0$ & $85.2 \pm 9.7$ \\
    $w/$ $fine$-$tuning$ & $\mathbf{95.5 \pm 4.3}$ & $\mathbf{93.6 \pm 4.1}$ & $\mathbf{93.4 \pm 5.3}$ & $\mathbf{87.4 \pm 8.1}$ \\
    \specialrule{1.5pt}{0pt}{0pt}
    \end{tabular}
    }
    \vspace{-15pt}
\end{table}

\subsection{Visualization of Adaptive Clip and Discussion}
\begin{figure}[htbp]
\centerline{\includegraphics[width=\columnwidth]{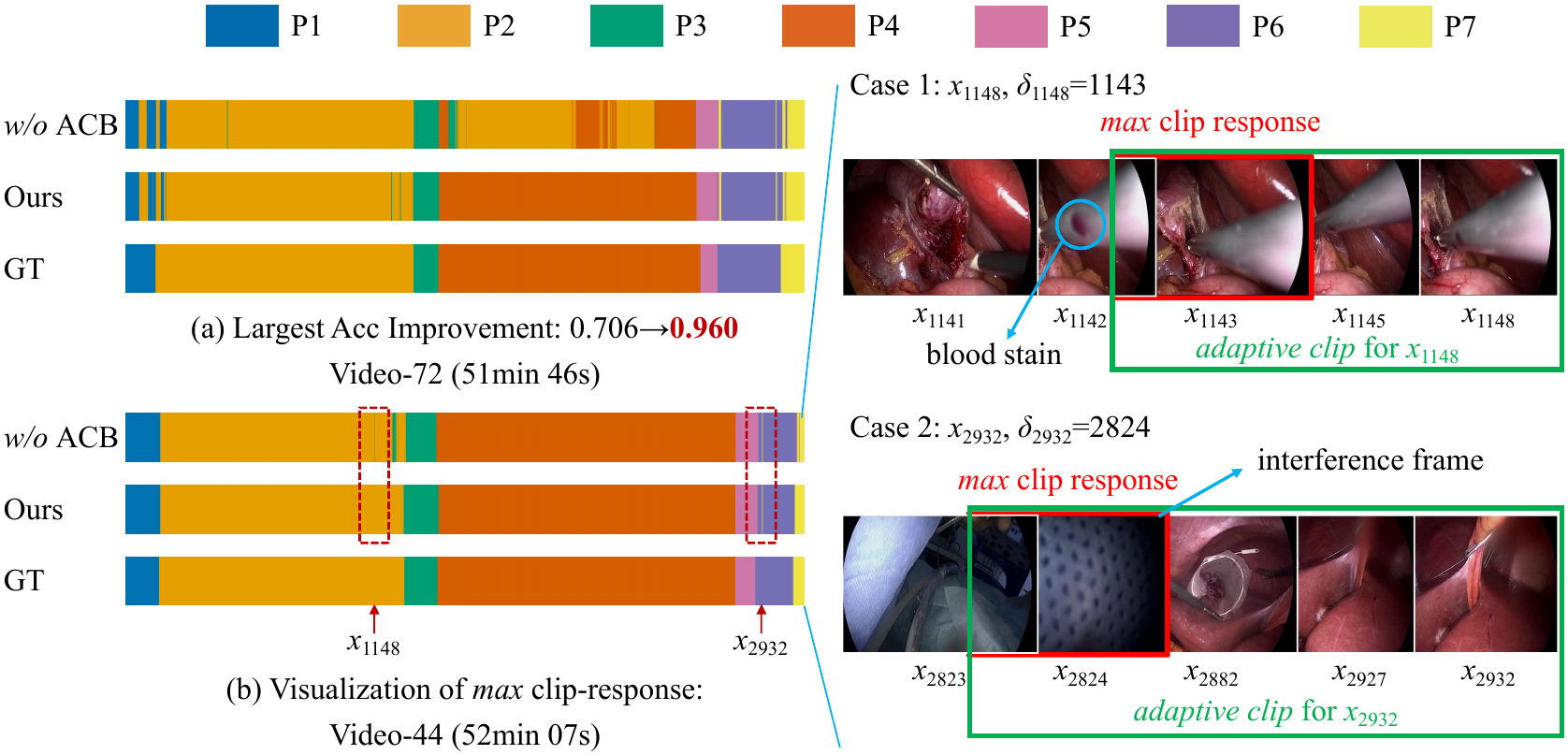}}
\caption{Visualization of adaptive clip. (a) shows the largest accuracy improvement compared with baseline ($w/o$ ACB). (b) visualizes the adaptive clip for two frames, a successful Case 1 and a failed Case 2.} 
\vspace{-5pt}
\label{fig4_bestclip}
\end{figure}

To better demonstrate our proposed ACB, we visualize and analyze a test case in detail. Fig.~\ref{fig4_bestclip} (a) showcases the video with the greatest accuracy improvement compared to the baseline ($w/o$ ACB), displaying smoother P4 recognition. 
In Fig.~\ref{fig4_bestclip} (b), we analyze $x_{1148}$ and $x_{2932}$, both of which are misclassified by the baseline, indicating that only using frame-wise features fails to correctly predict them. In Case 1, $\mathbf{AC}(1148)$ is $[x_{1143}$,~$x_{1148}]$, which can be observed that this clip is highly correlated with $x_{1148}$, indicating the Max-R accurately search for $\mathbf{AC}(1148)$. However, $x_{1142}$ contains blood stains, reducing its correlation with $x_{1148}$. Despite this, $\mathbf{AC}(1148)$ assists ACB in making the correct prediction. In Case 2, due to severe interference from irrelevant frame information, ACB also fails. 

Although Max-R utilizes \textit{suffix} sum to reduce sensitivity to noise and interference, there are still cases it cannot handle, such as Case 2 in Fig.~\ref{fig4_bestclip} (b). Therefore, in the future, we plan to further enhance the noise robustness of Max-R. Additionally, we will explore developing a mathematical model to statistically analyze $\mathbf{AC}(t)$ in surgical videos, incorporating this into Max-R design process to find $\mathbf{AC}(t)$ more efficiently. Moreover, we aim to investigate a more suitable fusion strategy for FWB and ACB, allowing to better leverage the information from two branches.

\section{Conclusion}

In this work, we propose a novel online surgical phase recognition network DACAT that combines frame-wise with adaptive clip-aware features to enhance the extraction capability of features associated with the current frame while reducing the introduction of additional interfering information and noise. We design a parameter-free max clip-response read-out mechanism that quickly identifies the clip most relevant to the current frame. 
Extensive experiments on Cholec80, M2CAI16, and AutoLaparo demonstrate that our DACAT surpasses existing SOTA methods, and achieves superior performance. Additionally, comprehensive ablation studies validate the rationality and effectiveness of our network. In the future, we plan to further enhance anti-interference ability of Max-R.



\vfill\pagebreak

\bibliographystyle{IEEEtran}
\bibliography{ref}

\begin{thebibliography}{10}
\providecommand{\url}[1]{#1}
\csname url@samestyle\endcsname
\providecommand{\newblock}{\relax}
\providecommand{\bibinfo}[2]{#2}
\providecommand{\BIBentrySTDinterwordspacing}{\spaceskip=0pt\relax}
\providecommand{\BIBentryALTinterwordstretchfactor}{4}
\providecommand{\BIBentryALTinterwordspacing}{\spaceskip=\fontdimen2\font plus
\BIBentryALTinterwordstretchfactor\fontdimen3\font minus \fontdimen4\font\relax}
\providecommand{\BIBforeignlanguage}[2]{{%
\expandafter\ifx\csname l@#1\endcsname\relax
\typeout{** WARNING: IEEEtran.bst: No hyphenation pattern has been}%
\typeout{** loaded for the language `#1'. Using the pattern for}%
\typeout{** the default language instead.}%
\else
\language=\csname l@#1\endcsname
\fi
#2}}
\providecommand{\BIBdecl}{\relax}
\BIBdecl

\bibitem{ref10resnet}
K.~He, X.~Zhang, S.~Ren, and J.~Sun, ``Deep residual learning for image recognition,'' in \emph{Proceedings of the IEEE conference on computer vision and pattern recognition}, 2016, pp. 770--778.

\bibitem{ref6Transformer}
A.~Vaswani, N.~Shazeer, N.~Parmar, J.~Uszkoreit, L.~Jones, A.~N. Gomez, L.~u. Kaiser, and I.~Polosukhin, ``Attention is all you need,'' in \emph{Advances in Neural Information Processing Systems}, vol.~30.\hskip 1em plus 0.5em minus 0.4em\relax Curran Associates, Inc., 2017, pp. 5998--6008.

\bibitem{ref7ViT}
A.~Dosovitskiy, L.~Beyer, A.~Kolesnikov, D.~Weissenborn, X.~Zhai, T.~Unterthiner, M.~Dehghani, M.~Minderer, G.~Heigold, S.~Gelly \emph{et~al.}, ``An image is worth 16x16 words: Transformers for image recognition at scale,'' in \emph{International Conference on Learning Representations}, 2020.

\bibitem{ref4LSTM}
S.~Hochreiter and J.~Schmidhuber, ``Long short-term memory,'' \emph{Neural computation}, vol.~9, no.~8, pp. 1735--1780, 1997.

\bibitem{ref5TCN}
C.~Lea, M.~D. Flynn, R.~Vidal, A.~Reiter, and G.~D. Hager, ``Temporal convolutional networks for action segmentation and detection,'' in \emph{proceedings of the IEEE Conference on Computer Vision and Pattern Recognition}, 2017, pp. 156--165.

\bibitem{ref11multi}
A.~P. Twinanda, S.~Shehata, D.~Mutter, J.~Marescaux, M.~De~Mathelin, and N.~Padoy, ``Endonet: a deep architecture for recognition tasks on laparoscopic videos,'' \emph{IEEE transactions on medical imaging}, vol.~36, no.~1, pp. 86--97, 2016.

\bibitem{ref12multi}
H.~Nakawala, R.~Bianchi, L.~E. Pescatori, O.~De~Cobelli, G.~Ferrigno, and E.~De~Momi, ```deep-onto' network for surgical workflow and context recognition,'' \emph{International journal of computer assisted radiology and surgery}, vol.~14, pp. 685--696, 2019.

\bibitem{ref13multimtrcnet}
Y.~Jin, H.~Li, Q.~Dou, H.~Chen, J.~Qin, C.-W. Fu, and P.-A. Heng, ``Multi-task recurrent convolutional network with correlation loss for surgical video analysis,'' \emph{Medical image analysis}, vol.~59, p. 101572, 2020.

\bibitem{ref14lstmSV-RCNet}
Y.~Jin, Q.~Dou, H.~Chen, L.~Yu, J.~Qin, C.-W. Fu, and P.-A. Heng, ``Sv-rcnet: workflow recognition from surgical videos using recurrent convolutional network,'' \emph{IEEE transactions on medical imaging}, vol.~37, no.~5, pp. 1114--1126, 2017.

\bibitem{ref15lstm}
F.~Yi and T.~Jiang, ``Hard frame detection and online mapping for surgical phase recognition,'' in \emph{International Conference on Medical Image Computing and Computer-Assisted Intervention}, vol. 11768.\hskip 1em plus 0.5em minus 0.4em\relax Springer, 2019, pp. 449--457.

\bibitem{ref16lstm}
X.~Gao, Y.~Jin, Q.~Dou, and P.-A. Heng, ``Automatic gesture recognition in robot-assisted surgery with reinforcement learning and tree search,'' in \emph{2020 IEEE international conference on robotics and automation (ICRA)}.\hskip 1em plus 0.5em minus 0.4em\relax IEEE, 2020, pp. 8440--8446.

\bibitem{ref17lstmTMRNet}
Y.~Jin, Y.~Long, C.~Chen, Z.~Zhao, Q.~Dou, and P.-A. Heng, ``Temporal memory relation network for workflow recognition from surgical video,'' \emph{IEEE Transactions on Medical Imaging}, vol.~40, no.~7, pp. 1911--1923, 2021.

\bibitem{ref18lstmBNPitfalls}
D.~Rivoir, I.~Funke, and S.~Speidel, ``On the pitfalls of batch normalization for end-to-end video learning: a study on surgical workflow analysis,'' \emph{Medical Image Analysis}, vol.~94, p. 103126, 2024.

\bibitem{ref19TCNTECNO}
T.~Czempiel, M.~Paschali, M.~Keicher, W.~Simson, H.~Feussner, S.~T. Kim, and N.~Navab, ``Tecno: Surgical phase recognition with multi-stage temporal convolutional networks,'' in \emph{International Conference on Medical Image Computing and Computer-Assisted Intervention}, vol. 12263.\hskip 1em plus 0.5em minus 0.4em\relax Springer, 2020, pp. 343--352.

\bibitem{ref20TransOpera}
T.~Czempiel, M.~Paschali, D.~Ostler, S.~T. Kim, B.~Busam, and N.~Navab, ``Opera: Attention-regularized transformers for surgical phase recognition,'' in \emph{International Conference on Medical Image Computing and Computer-Assisted Intervention}, vol. 12904.\hskip 1em plus 0.5em minus 0.4em\relax Springer, 2021, pp. 604--614.

\bibitem{ref21TransTransSVNet}
X.~Gao, Y.~Jin, Y.~Long, Q.~Dou, and P.-A. Heng, ``Trans-svnet: Accurate phase recognition from surgical videos via hybrid embedding aggregation transformer,'' in \emph{International Conference on Medical Image Computing and Computer-Assisted Intervention}, vol. 12904.\hskip 1em plus 0.5em minus 0.4em\relax Springer, 2021, pp. 593--603.

\bibitem{ref22Translovit}
Y.~Liu, M.~Boels, L.~C. Garcia-Peraza-Herrera, T.~Vercauteren, P.~Dasgupta, A.~Granados, and S.~Ourselin, ``Lovit: Long video transformer for surgical phase recognition,'' \emph{arXiv preprint arXiv:2305.08989}, 2023.

\bibitem{ref23TransCMTNet}
W.~Yue, H.~Liao, Y.~Xia, V.~Lam, J.~Luo, and Z.~Wang, ``Cascade multi-level transformer network for surgical workflow analysis,'' \emph{IEEE Transactions on Medical Imaging}, vol.~42, no.~10, pp. 2817--2831, 2023.

\bibitem{ref24TransSkit}
Y.~Liu, J.~Huo, J.~Peng, R.~Sparks, P.~Dasgupta, A.~Granados, and S.~Ourselin, ``Skit: a fast key information video transformer for online surgical phase recognition,'' in \emph{Proceedings of the IEEE/CVF International Conference on Computer Vision}, 2023, pp. 21\,074--21\,084.

\bibitem{ref25BN}
S.~Ioffe and C.~Szegedy, ``Batch normalization: Accelerating deep network training by reducing internal covariate shift,'' in \emph{International Conference on Machine Learning}.\hskip 1em plus 0.5em minus 0.4em\relax PMLR, 2015, pp. 448--456.

\bibitem{ref26convnext}
Z.~Liu, H.~Mao, C.-Y. Wu, C.~Feichtenhofer, T.~Darrell, and S.~Xie, ``A convnet for the 2020s,'' in \emph{Proceedings of the IEEE/CVF conference on computer vision and pattern recognition}, 2022, pp. 11\,976--11\,986.

\bibitem{ref27convnextv2}
S.~Woo, S.~Debnath, R.~Hu, X.~Chen, Z.~Liu, I.~S. Kweon, and S.~Xie, ``Convnext v2: Co-designing and scaling convnets with masked autoencoders,'' in \emph{Proceedings of the IEEE/CVF Conference on Computer Vision and Pattern Recognition}, 2023, pp. 16\,133--16\,142.

\bibitem{ref28spacememory}
S.~W. Oh, J.-Y. Lee, N.~Xu, and S.~J. Kim, ``Video object segmentation using space-time memory networks,'' in \emph{Proceedings of the IEEE/CVF international conference on computer vision}, 2019, pp. 9226--9235.

\bibitem{ref32uatd}
X.~Ding, X.~Yan, Z.~Wang, W.~Zhao, J.~Zhuang, X.~Xu, and X.~Li, ``Less is more: Surgical phase recognition from timestamp supervision,'' \emph{IEEE Transactions on Medical Imaging}, vol.~42, no.~6, pp. 1897--1910, 2023.

\bibitem{ref33sphase}
J.~Long, J.~Hong, Z.~Wang, T.~Chen, Y.~Chen, and L.~Yang, ``Sphase: Multi-modal and multi-branch surgical phase segmentation framework based on temporal convolutional network,'' in \emph{2023 IEEE International Conference on Bioinformatics and Biomedicine (BIBM)}.\hskip 1em plus 0.5em minus 0.4em\relax IEEE, 2023, pp. 586--593.

\bibitem{ref29Cholec80}
A.~P. Twinanda, S.~Shehata, D.~Mutter, J.~Marescaux, M.~De~Mathelin, and N.~Padoy, ``Endonet: a deep architecture for recognition tasks on laparoscopic videos,'' \emph{IEEE transactions on medical imaging}, vol.~36, no.~1, pp. 86--97, 2016.

\bibitem{ref30m2cai16}
\BIBentryALTinterwordspacing
A.~P. Twinanda, S.~Shehata, D.~Mutter, J.~Marescaux, M.~De~Mathelin, and N.~Padoy. (2016) Workshop and challenges on modeling and monitoring of computer assisted interventions. [Online]. Available: \url{http://camma.u-strasbg.fr/m2cai2016/}
\BIBentrySTDinterwordspacing

\bibitem{ref31autolaparo}
Z.~Wang, B.~Lu, Y.~Long, F.~Zhong, T.-H. Cheung, Q.~Dou, and Y.~Liu, ``Autolaparo: A new dataset of integrated multi-tasks for image-guided surgical automation in laparoscopic hysterectomy,'' in \emph{International Conference on Medical Image Computing and Computer-Assisted Intervention}, vol. 13437.\hskip 1em plus 0.5em minus 0.4em\relax Springer, 2022, pp. 486--496.

\end{thebibliography}

\end{document}